\documentclass{article}
\usepackage{ijcai19}

\usepackage{times}
\usepackage{soul}
\usepackage{url}
\usepackage[switch]{lineno}
\usepackage[hidelinks]{hyperref}
\usepackage[utf8]{inputenc}
\usepackage[small]{caption}
\usepackage{graphicx}
\usepackage{amsmath}
\usepackage{booktabs}
\usepackage{comment}
\usepackage{multirow}
\usepackage{tabularx}
\usepackage{amssymb}
\usepackage{bm}

\usepackage{xspace}

\title{EmotionX-IDEA: Emotion BERT -- an Affectional Model for Conversation}

\author{
Yen-Hao Huang\and
Ssu-Rui Lee\and
Mau-Yun Ma\and
Yi-Hsin Chen\and
Ya-Wen Yu\and
Yi-Shin Chen\footnote{Corresponding Author}\\
\affiliations
Intelligent Data Engineering and Applications Laboratory\\
National Tsing Hua University\\
Hsinchu, Taiwan\\
\emails
\{yenhao0218, gn01697933, brian41005, eunicebes, evan800112, yishin\}@gmail.com
}

\begin{document}

\maketitle

\begin{abstract}

In this paper, we investigate the emotion recognition ability of the pre-training language model, namely BERT. By the nature of the framework of BERT, a two sentence structure, we adapt BERT to continues dialogue emotion prediction tasks, which rely heavily on the sentence-level context-aware understanding. The experiments show that by mapping the continues dialogue into a causal utterance pair, which is constructed by the utterance and the reply utterance, models can better capture the emotions of the reply utterance. The present method have achieved $0.815$ and $0.885$ micro F1 score in the testing dataset of \textit{Friends} and \textit{EmotionPush}, respectively.

\end{abstract}


\section{Introduction}

Emotion detection has long been a topic of interest to scholars in natural language processing (NLP) domain.  
Researchers aim to recognize the emotion behind the text and distribute similar ones into the same group. 
Establishing an emotion classifier can not only understand each user's feeling but also be extended to various application, for example, the motivation behind a user's interests~\cite{elvis2015interest}.
Based on releasing of large text corpus on social media and the emotion categories proposed by ~\cite{ekman1987universals,plutchik2001nature}, numerous models have provided and achieved fabulous precision so far. For example, DeepMoji~\cite{felbo2017using} which utilized transfer learning concept to enhance emotions and sarcasm understanding behind the target sentence. CARER~\cite{saravia2018carer} learned contextualized affect representations to make itself more sensitive to rare words and the scenario behind the texts.

As methods become mature, text-based emotion detecting applications can be extended from a single utterance to a dialogue contributed by a series of utterances. Table~\ref{tab:sentence_dialogue} illustrates the difference between single utterance and dialogue emotion recognition. The same utterances in Table~\ref{tab:sentence_dialogue}, even the same person said the same sentence, the emotion it convey may be various, which may depend on different background of the conversation, tone of speaking or personality. Therefore, for emotion detection, the information from preceding utterances in a conversation is relatively critical.

\begin{table}[!ht]
    \caption{Emotions depending on the context}
    \label{tab:sentence_dialogue}
    \footnotesize
    \centering
    \begin{tabular}{rl}
    \toprule
    \textbf{Monica} & I'm gonna miss you! \\ 
    \textbf{Rachel} & I mean it's the end of an era! \\
    \textbf{Monica} &  \textit{\textbf{I know! (sadness)}}\\
    \midrule
    \textbf{Chandler} & So, what do you think? \\ 
    \textbf{Ross} & I think It's the most beautiful table  \\
    {} & I've ever seen. \\
    \textbf{Chandler} & \textit{\textbf{I know! (joy)}}\\
    \midrule
    \textbf{Monica} & Now, this is last minute so I want to  \\
    {} & apologize for the mess. Okay? \\ 
    \textbf{Rachel} & Oh my God! It sure didn't look this way\\ {} & when I lived here. \\
    \textbf{Monica} & \textit{\textbf{I know! (surprise)}}\\
    \toprule
    \end{tabular}
\end{table}

In SocialNLP 2019 EmotionX, the challenge is to recognize emotions for all utterances in \textit{\textbf{EmotionLines}} dataset, a dataset consists of dialogues. According to the needs for considering context at the same time, we develop two classification models, inspired by bidirectional encoder representations from transformers ({BERT})~\cite{devlin2018bert}, \textbf{FriendsBERT} and \textbf{ChatBERT}. In this paper, we introduce our approaches including causal utterance modeling, model pre-training, and fine-turning.

\section{Dataset}

\textit{\textbf{EmotionLines}}~\cite{chen2018emotionlines} is a dialogue dataset composed of two subsets, \textit{\textbf{Friends}} and \textit{\textbf{EmotionPush}}, according to the source of the dialogues. The former comes from the scripts of the Friends TV sitcom. The other is made up of Facebook messenger chats. Each subset includes $1,000$ English dialogues, and each dialogue can be further divided into a few consecutive utterances. All the utterances are annotated by five annotators on a crowd-sourcing platform (Amazon Mechanical Turk), and the labeling work is only based on the textual content. Annotator votes for one of the seven emotions, namely Ekman’s six basic emotions~\cite{ekman1987universals}, plus the neutral. If none of the emotion gets more than three votes, the utterance will be marked as ``non-neutral”.

For the datasets, there are properties worth additional mentioning. Although \textit{\textbf{Friends}} and \textit{\textbf{EmotionPush}} share the same data format, they are quite different in nature. \textit{\textbf{Friends}} is a speech-based dataset which is annotated dialogues from the TV sitcom. It means most of the utterances are generated by the a few main characters. The personality of a character often affects the way of speaking, and therefore ``who is the speaker" might provide extra clues for emotion prediction. In contrast, \textit{\textbf{EmotionPush}} does not have this trait due to the anonymous mechanism. In addition, features such as typo, hyperlink, and emoji that only appear in chat-based data will need some domain-specific techniques to process.

Incidentally, the objective of the challenge is to predict the emotion for each utterance. Just, according to EmotionX 2019 specification, there are only four emotions be selected as our label candidates, which are \textit{Joy}, \textit{Sadness}, \textit{Anger}, and \textit{Neutral}. These emotions will be considered during performance evaluation. The technical detail will also be introduced and discussed in following Section~\ref{sub:data_preprocessing} and Section~\ref{sub:exp_data}.

\begin{figure*}
    \centering
    \includegraphics[width=1\linewidth]{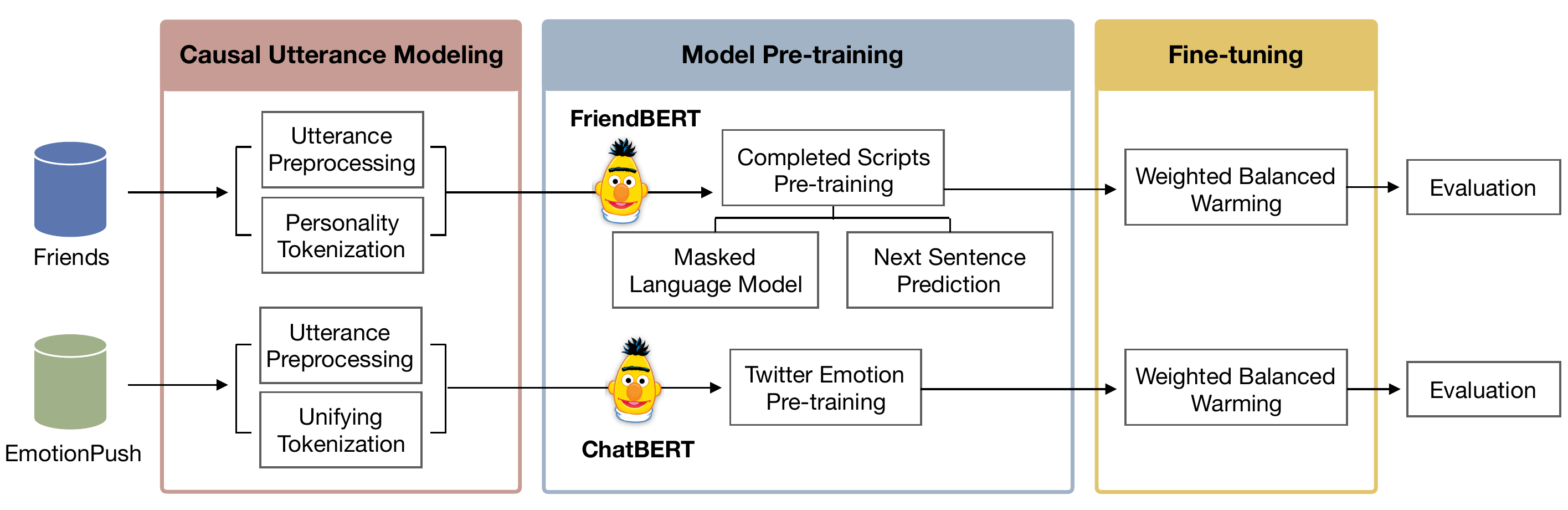}
    \caption{Framework}
    \label{fig:Framework}
\end{figure*}

\section{Model Description}

For this challenge, we adapt BERT which is proposed by \cite{devlin2018bert} to help understand the context at the same time. Technically, BERT, designed on end-to-end architecture, is a deep pre-trained transformer encoder that dynamically provides language representation and BERT already achieved multiple state-of-the-art results on GLUE benchmark \cite{DBLP:journals/corr/abs-1804-07461} and many tasks. 
A quick recap for BERT's architecture and its pre-training tasks will be illustrated in the following subsections.

\subsection{Model Architecture}

BERT, the Bidirectional Encoder Representations from Transformers, consists of several transformer encoder layers that enable the model to extract very deep language features on both token-level and sentence-level. Each transformer encoder contains multi-head self-attention layers that provide ability to learn multiple attention feature of each word from their bidirectional context. The transformer and its self-attention mechanism are proposed by \cite{vaswani2017attention}. 
This self-attention mechanism can be interpreted as a key-value mapping given query. 
By given the embedding vector for token input, the query ($Q$), key ($K$) and value ($V$) are produced by the projection from each three parameter matrices where \(W^Q \in \mathbb{R}^{d_{{\rm model}} \times d_{k}}, W^K \in \mathbb{R}^{d_{\rm model} \times d_{k}}\) and \(W^V \in \mathbb{R}^{d_{\rm model} \times d_{v}}\). 
The self-attention~\cite{vaswani2017attention} is formally represented  as:

\begin{equation}
{\rm Attention}(Q, K, V) = {\rm softmax} (\frac{QK^T}{\sqrt{d_k}})V
\end{equation}

The \( d_k = d_v = d_{\rm model} = 1024\) in BERT large version and 768 in BERT base version. Once model can extract attention feature, we can extend one self-attention into multi-head self-attention, this extension makes sub-space features can be extracted in same time by this multi-head configuration. 
Overall, the multi-attention mechanism is adopt for each transformer encoder, and several of encoder layer will be stacked together to form a deep transformer encoder.


For the model input, BERT allow us take one sentence as input sequence or two sentences together as one input sequence, and the maximum length of input sequence is $512$. The way that BERT was designed is for giving model the sentence-level and token-level understanding. In two sentences case, a special token ({\tt [SEP]}) will be inserted between two sentences. In addition, the first input token is also a special token ({\tt [CLS]}), and its corresponding ouput will be vector place for classification during fine-tuning. 
The outputs of the last encoder layer corresponding to each input token can be treated as word representations for each token, and the word representation of the first token ({\tt [CLS]}) will be consider as classification (output) representation for further fine-tuning tasks.
In BERT, this vector is denoted as \(C \in \mathbb{R}^{d_{\rm model}} \), and a classification layer is denoted as \( W \in \mathbb{R}^{K \times d_{\rm model}}\), where $K$ is number of classification labels. Finally, the prediction $P$ of BERT is represented as \(P = {\rm softmax}(CW^T)\).

\subsection{Pre-training Tasks}

In pre-training, intead of using unidirectional language models, BERT developed two pre-training tasks: (1) Masked LM (cloze test) and (2) Next Sentence Prediction. At the first pre-training task, bidirectional language modeling can be done at this cloze-like pre-training. In detail, 15\% tokens of input sequence will be masked at random and model need to predict those masked tokens. The encoder will try to learn contextual representations from every given tokens due to masking tokens at random. Model will not know which part of the input is going to be masked, so that the information of each masked tokens should be inferred by remaining tokens. At Next Sentence Prediction, two sentences concatenated together will be considered as model input. In order to give model a good nature language understanding, knowing relationship between sentence is one of important abilities. When generating input sequences, 50\% of time the sentence B is actually followed by sentence A, and rest 50\% of the time the sentence B will be picked randomly from dataset, and model need to predict if the sentence B is next sentence of sentence A. That is, the attention information will be shared between sentences. Such sentence-level understanding may have difficulties to be learned at first pre-training task (Masked LM), therefore, the pre-training task (NSP) is developed as second training goal to capture the cross sentence relationship.

In this competition, limited by the size of dataset and the challenge in contextual emotion recognition, we consider BERT with both two pre-training tasks can give a good starting point to extract emotion changing during dialogue-like conversation. Especially the second pre-training task, it might be more important for dialogue-like conversation where the emotion may various by the context of continuous utterances. That is, given a set of continues conversations, the emotion of current utterance might be influenced by previous utterance. By this assumption and with supporting from the experiment results of BERT, we can take sentence A as one-sentence context and consider sentence B as the target sentence for emotion prediction. The detail will be described in Section~\ref{sec:method}.

\section{Methodology}
\label{sec:method}

The main goal of the present work is to predict the emotion of utterance within the dialogue.
Following are four major difficulties we concern about:
\begin{enumerate}
    \item The emotion of the utterances depends not only on the text but also on the interaction happened earlier.
    \item The source of the two datasets are different. \textit{Friends} is speech-based dialogues and \textit{EmotionPush} is chat-based dialogues. It makes datasets possess different characteristics.
    \item There are only $1,000$ dialogues in both training datasets which are not large enough for the stability of training a complex neural-based model.
    \item The prediction targets (emotion labels) are highly unbalanced. 
\end{enumerate}
The proposed approach is summarized in Figure~\ref{fig:Framework}, which aims to overcome these challenges. The framework could be separated into three steps and described as follow:

\begin{table*} [!ht]
    \caption{An example of sentence representation}
    \label{tab:sentence_representation}
    \footnotesize
    \centering
    \begin{tabular}{lll|ll}
    \toprule
    \textbf{Speaker} & \textbf{Utterance} & \textbf{Emotion} & \textbf{Representation} & \textbf{Label}\\  
    \midrule
    Joey & What?! & surprise & \text{\tt {\scriptsize [CLS] What{?!}\;[SEP] [None] [SEP]}} &  \text{\tt {\scriptsize surprise}} \\
    Chandler & What's wrong with you? & non-neutral & \text{\tt {\scriptsize [CLS] What{'}s wrong with you{?}\;[SEP] What{?!}\;[SEP]}} & \text{\tt {\scriptsize non-neutral}} \\ 
    Joey & Nothing! & neutral & \text{\tt {\scriptsize [CLS] Nothing{!}\;[SEP] What{'}s wrong with you{?}\;[SEP]}} & \text{\tt {\scriptsize neutral}} \\
    \toprule
    \end{tabular}
\end{table*}

\begin{table*} [!ht]
    \caption{An example of personality tokenization}
    \label{tab:personality_tokenization}
    \footnotesize
    \centering
    \renewcommand{\tabularxcolumn}{m} 
    \begin{tabularx}{\textwidth}{lll| X l}
    \toprule
    \textbf{Speaker} & \textbf{Utterance} & \textbf{Emotion} & \textbf{Representation (with personality tokenization)} & \textbf{Label} 
    \tabularnewline
    \midrule
    Janice & I'm sorry. & sadness & 
    \text{\tt {\scriptsize [CLS] I'm sorry.\;[SEP] [None] [SEP]}} & \text{\tt {\scriptsize sadness}} 
    \tabularnewline
    Chandler & Ohhh. Don't go. & sadness & 
    \text{\tt {\scriptsize [CLS] [Chandler] [says] Ohhh. Don't go.\;[SEP]}}
    \text{\tt {\scriptsize I'm sorry.\;[SEP]}} & \text{\tt {\scriptsize sadness}}  \tabularnewline 
    Janice & No, I gotta go. & non-neutral & 
    \text{\tt {\scriptsize [CLS] No, I gotta go.\;[SEP]}} 
    \text{\tt {\scriptsize [Chandler] [says] Ohhh.Don't go.\;[SEP]}} & \text{\tt {\scriptsize non-neutral}} \tabularnewline
    \bottomrule
    \end{tabularx}
\end{table*}

\subsection{Causal Utterance Modeling}
\label{sub:data_preprocessing}

Given a dialogue $D^{(i)}$ which includes sequence of utterances denoted as $D^{(i)}=(u^{(i)}_{1}, u^{(i)}_{2}, ..., u^{(i)}_{n})$, where $i$ is the index in dataset and $n$ is the number of utterances in the given dialogue. In order to conserve the emotional information of both utterance and conversation, we rearrange each two consecutive utterances $u_{t}, u_{t-1}$ into a single sentence representation $x_{t}$ as 
\begin{equation}
x^{(i)}_{t} = {\rm concat}(u^{(i)}_{t}, u^{(i)}_{t-1})
\label{equ:sentence_representation}
\end{equation}
The corresponding sentence representation corpus $X^{(i)}$ are denoted as $X^{(i)}=(x^{(i)}_{1}, x^{(i)}_{2}, ..., x^{(i)}_{n})$. 
Note that the first utterance within a conversation does not have its causal utterance (previous sentence), therefore, the causal utterance will be set as {\tt [None]}.
A practical example of sentence representation is shown in Table~\ref{tab:sentence_representation}.

Since the characteristics of two datasets are not identical, we customize different causal utterance modeling strategies to refine the information in text.

For \textbf{\textit{Friends}}, there are two specific properties. The first one is that most dialogues are surrounding with the six main characters, including \textit{Rachel}, \textit{Monica}, \textit{Phoebe}, \textit{Joey}, \textit{Chandler}, and \textit{Ross}. The utterance ratio of given by the six roles is up to $83.4\%$. 
Second, the personal characteristics of the six characters are very clear. Each leading role has its own emotion undulated rule.
To make use of these features, we introduce the \textbf{personality tokenization} which help learning the personality of the six characters. Personality tokenization concatenate the \textit{speaker} and \textit{says} tokens before the input \textit{utterance} if the speaker is one of the six characters. The example is shown in Table~\ref{tab:personality_tokenization}.

For \textbf{\textit{EmotionPush}}, the text are informal chats which including like slang, acronym, typo, hyperlink, and emoji. Another characteristic is that the specific name entities are tokenized with random index. (e.g. ``organization\_80'', ``person\_01'', and ``time\_12'').
We consider some of these informal text are related to expressing emotion such as repeated typing, purposed capitalization, and emoji (e.g. ``:D'', ``:('', and ``\textless 3'')). Therefore, we keep most informal expressions but only process hyperlinks, empty utterance, and name entities by unifying the tokens.

\subsection{Model Pre-training}

Since the size of both datasets are not large enough for complex neural-based model training as well as BERT model is only pre-train on formal text datasets, the issues of overfitting and domain bias are important considerations for design the pre-training process.

To avoid our model overfitting on the training data and increase the understanding of informal text, we adapted BERT and derived two models, namely \textbf{FriendsBERT} and \textbf{ChatBERT}, with different pre-training tasks before the formal training process for \textit{Friends} and \textit{EmotionPush} dataset, respectively. 
The pre-training strategies are described below.


For pre-training \textbf{FriendsBERT}, we collect the completed scripts of all ten seasons of \textit{Friends} TV shows from emorynlp\footnote{http://nlp.mathcs.emory.edu} which includes 3,107 scenes within 61,309 utterances. 
All the utterances are followed the preprocessing methods mentions above to compose the corpus for Masked language model pre-training task. The consequent utterances in the same scenes are considered as the consequent sentences to pre-train the Next Sentence Prediction task. In the pre-training process, the training loss is the sum of the mean likelihood of two pre-train tasks.

For pre-training \textbf{ChatBERT}, we pre-train our model on the Twitter dataset, since the text and writing style on Twitter are close to the chat text where both may involved with many informal words or emoticons as well. The Twitter emotion dataset, 8 basic emotions from \textit{emotion wheel}~\cite{ekman1987universals}, was collected by \textit{twitter streaming API} with specific emotion-related hashtags, such as \textit{\#anger, \#joy, \#cry, \#sad} and etc. The hashtags in tweets are treated as emotion label for model fine-tuning.
The tweets were fine-grined processing followed the rules in~\cite{abdul2017emonet,saravia2018carer}, including duplicate tweets removing, the emotion hashtags must appearing in the last position of a tweet, and etc.
The statis of tweets were summarized in Table~\ref{tab:twitter_statistics}.
Each tweet and corresponding emotion label composes an emotion classification dataset for pre-training. 

\begin{table}[]

    \caption{Statistics for Twitter Dataset}
    \label{tab:twitter_statistics}
    \centering
    \begin{tabular}{c|c|c}
        \toprule
        \textbf{Emotions} & \textbf{Amount} & \textbf{Hashtags}\\
        \hline\hline
        \textbf{Anger} & 102,289 & \#mad, \#pissed \\
        \textbf{Anticipation} & 3,975 & \#pumped, \#ready \\
        \textbf{Disgust} & 8,934 & \#awful, \#eww \\
        \textbf{Fear} & 102,468 & \#fear, \#worried \\
        \textbf{Joy} & 167,027 & \#fun, \#joy \\
        \textbf{Sadness} & 214,454 & \#depressed, \#grief \\
        \textbf{Surprise} & 46,101 & \#strange, \#surprise \\
        \textbf{Trust} & 19,222 & \#hope, \#secure \\
        \bottomrule
    \end{tabular}
\end{table}{}


\begin{table}
    \caption{Emotions Distribution of two dataset}
    \label{tab:exp_data}
    \footnotesize
    \centering
    \begin{tabular}{lll}
    \toprule
    \multicolumn{1}{c}{\multirow{2}{*}{\textbf{Processing}}} & \multicolumn{2}{c}{\textbf{EmotionLines}} \\
    \cmidrule(lr){2-3}
    \multicolumn{1}{c}{} & \textbf{Friends} & \textbf{EmotionPush} \\
    \midrule
    \textbf{Dialogue} & 1,000 & 1,000 \\
    \cmidrule(lr){2-3}
    \textbf{Utterance} & 14,503 & 14,742 \\
    \cmidrule(lr){2-3}
    \textbf{Utterance (filtered)} & 9,479 & 12,609 \\
    \cmidrule(lr){2-3}
    \textbf{Train / Val} & 7,660 / 1,837 & 10,145 / 2,464 \\
    \midrule
    \multicolumn{3}{c}{\textbf{Emotions in Training / Validation set}} \\
    \midrule
    Anger & 598 / 161 & 103 / 37 \\
    Joy & 1,406 / 304 & 1,642 / 458 \\
    Neutral & 5,243 / 1,287 & 7,973 / 1,882 \\
    Sadness & 413 / 85 & 427 / 87\\
    \bottomrule
    \end{tabular} 
\end{table}

\begin{table*} [ht]
    \caption{Validation Results (\textit{Friends})}    
    \label{tab:compare_friends}
    \footnotesize
    \centering
    \begin{tabular}{c|c|c|c|c|c|c|c|c|c|c|c|c|c}
    \toprule
     & \multicolumn{3}{c}{\textbf{Anger}} & \multicolumn{3}{|c}{\textbf{Joy}}  & \multicolumn{3}{|c}{\textbf{Neutral}}  & \multicolumn{3}{|c}{\textbf{Sadness}} & \multicolumn{1}{|c}{\textbf{Overall}} \\
     \hline
     Models & P. & R. & F1 & P. & R. & F1 & P. & R. & F1 & P. & R. & F1 & F1 \\
     \hline\hline
     \textbf{BOW-LR} & $0.65$ & $0.31$ & $0.42$ & $0.66$ & $0.66$ & $0.66$ & $0.85$ & $0.94$ & $0.89$ & $0.51$ & $0.21$ & $0.30$ & $0.80$ \\
     \textbf{BOW-RF} & $0.74$ & $0.30$ & $0.43$ & $0.66$ & $0.62$ & $0.64$ & $0.84$ & $0.96$ & $0.89$ & $0.83$ & $0.18$ & $0.29$ & $0.81$ \\
     \textbf{TFIDF-RF} & $0.61$ & $0.30$ & $0.40$ & $0.62$ & $0.65$ & $0.63$ & $0.86$ & $0.94$ & $0.90$ & $0.53$ & $0.19$ & $0.28$ & $0.80$ \\
     \textbf{TextCNN} & $0.74$ & $0.40$ & $0.52$ & $0.65$ & $0.71$ & $0.68$ & $0.87$ & $0.94$ & $0.90$ & $0.70$ & $0.25$ & $0.37$ & $0.82$ \\
     \textbf{C-TextCNN} & $0.71$ & $0.49$ & $0.58$ & $0.70$ & $0.70$ & $0.70$ & $0.87$ & $0.94$ & $0.90$ & $0.62$ & $0.25$ & $0.35$ & $0.83$ \\
     \hline
     \textbf{FriendsBERT-{base}-s} & $0.76$ & $0.58$ & $0.66$ & $0.79$ & $0.71$ & $0.75$ & $0.88$ & $0.95$ & $0.91$ & $0.61$ & $0.36$ & $0.46$ & $0.84$ \\     
     \textbf{FriendsBERT-{base}} & $0.78$ & $0.58$ & $0.66$ & $0.74$ & $0.78$ & $0.76$ & $0.89$ & $0.93$ & $0.91$ & $0.65$ & $0.38$ & $0.48$ & $0.85$ \\
     \textbf{FriendsBERT-{large}} & $0.80$ & $0.61$ & $0.69$ & $0.76$ & $0.83$ & $0.79$ & $0.91$ & $0.93$ & $0.92$ & $0.64$ & $0.41$ & $0.50$ & $0.86$ \\
     \bottomrule
    \end{tabular}
\end{table*}

\subsection{Fine-tuning}

Since our emotion recognition task is treated as a sequence-level classification task, the model would be fine-tuned on the processed training data.
Following the BERT construction, we take the first embedding vector which corresponds to the special token {\tt [CLS]} from the final hidden state of the Transformer encoder. 
This vector represents the embedding vector of the corresponding conversation utterances which is denoted as $\bm{C} \in \mathbb{R}^{H}$, where $H$ is the embedding size. 
A dense neural layer is treated as a classification layer which consists of parameters $\bm{W} \in \mathbb{R}^{K\times H}$ and $\bm{b} \in \mathbb{R}^{K}$, where $K$ is the number of emotion class. 
The emotion prediction probabilities $\bm{P} \in \mathbb{R}^{K}$ are computed by a softmax activation function as
\begin{equation}
\bm{P} = {\rm softmax}(\bm{C}\bm{W}^T+\bm{b})
\label{equ:prediction}
\end{equation}
All the parameters in BERT and the classification layer would be fine-turned together to minimize the Negative Log Likelihood (NLL) loss function, as Equation~(\ref{equ:loss}), based on the ground truth emotion label $c$.
\begin{equation}
\mathcal{L}=-\frac{1}{N} \sum_{i=1}^{N} \log \left(\hat{p}^{(i)}_{c}\right)
\label{equ:loss}
\end{equation}

In order to tackle the problem of highly unbalanced emotion labels, we apply weighted balanced warming on NLL loss function, as Equation~(\ref{equ:weightedloss}), in the first epoch of fine-tuning procedure.
\begin{equation}
\mathcal{L}=-\frac{1}{\sum_{i=1}^{N} w_{c}^{(i)}} \sum_{i=1}^{N} \log \left(w_{c}\hat{p}^{(i)}_{c}\right)
\label{equ:weightedloss}
\end{equation}
where $\bm{w}$ are the weights of corresponding emotion label $c$ which are computed and normalize by the frequency as
\begin{equation}
w_{c} = \frac{{\rm min}({\rm freq}(\bm{c}))}{{\rm freq}(c)},\; \forall c \in \bm{c}
\label{equ:weightedcount}
\end{equation}

By adding the weighted balanced warming on NLL loss, the model could learn to predict the minor emotions (e.g. \textit{anger} and \textit{sadness}) earlier and make the training process more stable.
Since the major evaluation metrics micro F1-score is effect by the number of each label, we only apply the weighted balanced warming in first epoch to optimize the performance.

\begin{table} [ht]
    \caption{Experimental Setup of Proposed Model}
    \label{tab:our_setting}
    \footnotesize
    \centering
    \begin{tabular}{c|c|c}
    \toprule
    \textbf{} & \textbf{FriendsBERT} & \textbf{ChatBERT} \\ 
    \hline \hline
    \textbf{Pre-trained weights} & BERT-uncased & BERT-cased \\ 
    \textbf{Batch size} & $8$ & $4$ \\ 
    \textbf{Learning rate (Adam)} & $2.5 \times 10^{-6}$ & $2.5 \times 10^{-6}$ \\ 
    \textbf{Number of epochs} & $3$ & $2$ \\ 
    \hline
    \textbf{Max length (input tokens)} & $113$ & $249$ \\ 
    \textbf{Dropout rate (last layer)} & $0.75$ & $0.75$ \\ 
    \bottomrule
    \end{tabular}
\end{table}

\section{Experiments}

Since the EmotionX challenge only provided the gold labels in training data, we pick the best performance model (weights) to predict the testing data. In this section, we present the experiment and evaluation results.

\subsection{Experimental Setup}
\label{sub:exp_data}

The EmotionX challenge consists of $1,000$ dialogues for both \textit{Friends} and \textit{EmotionPush}. In all of our experiments, each dataset is separated into top $800$ dialogues for training and last $200$ dialogues for validation. Since the EmotionX challenge considers only the four emotions (\textit{anger}, \textit{joy}, \textit{neutral}, and \textit{sadness}) in the evaluation stage, we ignore all the data point corresponding to other emotions directly. The details of emotions distribution are shown in Table~\ref{tab:exp_data}.

The hyperparameters and training setup of our models (\textbf{FriendsBERT} and \textbf{ChatBERT}) are shown in Table~\ref{tab:our_setting}. Some common and easily implemented methods are selected as the baselines embedding methods and classification models. The baseline embedding methods are including \textit{bag-of-words} (\textbf{BOW}), \textit{term frequency–inverse document frequency} (\textbf{TFIDF}), and neural-based word embedding. The classification models are including Logistic Regression (LR), Random Forest (RF), \textbf{TextCNN}~\cite{kim2014convolutional} with initial word embedding as GloVe~\cite{pennington2014glove}, and our proposed model. All the experiment results are based on the best performances of validation results.

\subsection{Performance}
The experiment results of validation on \textit{Friends} are shown in Table~\ref{tab:compare_friends}. The proposed model and baselines are evaluated based on the Precision (P.), Recall (R.), and F1-measure (F1).

For the traditional baselines, namely \textbf{BOW} and \textbf{TFIDF}, we observe that they achieve surprising high F1 scores around $0.81$, however, the scores for \textit{Anger} and \textit{Sadness} are lower. This explains that traditional approaches tend to predict the labels with large sample size, such as \textit{Joy} and \textit{Neutral}, but fail to take of scarce samples even when an ensemble \textit{random forest} classifier is adopted. In order to prevent the unbalanced learning, we choose the weighted loss mechanism for both \textbf{TextCNN} and causal modeling TextCNN (\textbf{C-TextCNN}), these models suffer less than the traditional baselines and achieve a slightly balance performance, where there are around $15$\% and $7$\% improvement on \textit{Anger} and \textit{Sadness}, respectively. 
We following adopt the casual utterance modeling to original \textbf{TextCNN}, mapping previous utterance as well as target utterance into model. 
The causal utterance modeling improve the \textbf{C-TextCNN} over \textbf{TextCNN} for $6$\%, $2$\% and $1$\% on \textit{Anger}, \textit{Joy} and overall F1 score. 
Motivated from these preliminary experiments, the proposed \textbf{FriendsBERT} also adopt the ideas of both weighted loss and causal utterance modeling. 
As compared to the original BERT, single sentence BERT (\textbf{FriendsBERT-base-s}), the proposed \textbf{FriendsBERT-base} improve $1$\% for \textit{Joy} and overall F1, and $2$\% for \textit{Sadness}. For the final validation performance, our proposed approach achieves the highest scores, which are $0.85$ and $0.86$ for \textbf{FriendsBERT-base} and \textbf{FriendsBERT-large}, respectively.

Overall, the proposed \textbf{FriendsBERT} successfully captures the sentence-level context-awarded information and outperforms all the baselines, which not only achieves high performance on large sample labels, but also on small sample labels. The similar settings are also adapted to \textit{EmotionPush} dataset for the final evaluation. 

\subsection{Evaluation Results}

The testing dataset consists of $240$ dialogues including $3,296$ and $3,536$ utterances in \textit{Friends} and \textit{EmotionPush} respectively. We re-train our \textbf{FriendsBERT} and \textbf{ChatBERT} with top $920$ training dialogues and predict the evaluation results using the model performing the best validation results. The results are shown in Table~\ref{tab:eval_friends} and Table~\ref{tab:eval_emotionpush}. The present method achieves $81.5\%$ and $88.5\%$ micro F1-score on the testing dataset of \textit{Friends} and \textit{EmotionPush}, respectively.

\begin{table} [ht]
    \caption{Evaluation (Testing) Results of \textit{Friends}}
    \label{tab:eval_friends}
    \footnotesize
    \centering
    \begin{tabular}{c|c|c|c|c}
    \toprule
    \textbf{} & \textbf{precision} & \textbf{recall} & \textbf{f1-score} & \textbf{support} \\ 
    \hline \hline
    \textbf{Anger} & $0.716$ & $0.681$ & $0.698$ & $141$ \\ 
    \textbf{Joy} & $0.875$ & $0.663$ & $0.755$ & $505$ \\ 
    \textbf{Neutral} & $0.814$ & $0.942$ & $0.873$ & $1,035$ \\ 
    \textbf{Sadness} & $0.713$ & $0.512$ & $0.596$ & $121$ \\ 
    \hline \hline
    \textbf{Micro AVG} & $0.815$ & $0.815$ & $0.815$ & $1,802$ \\ 
    \textbf{Macro AVG} & $0.779$ & $0.700$ & $0.731$ & $1,802$ \\ 
    \textbf{Weighted AVG} & $0.816$ & $0.815$ & $0.808$ & $1,802$ \\ 
    \bottomrule
    \end{tabular}
\end{table}

\begin{table} [ht]
    \caption{Evaluation (Testing) Results of \textit{EmotionPush}}
    \label{tab:eval_emotionpush}
    \footnotesize
    \centering
    \begin{tabular}{c|c|c|c|c}
    \toprule
    \textbf{} & \textbf{precision} & \textbf{recall} & \textbf{f1-score} & \textbf{support} \\ 
    \hline \hline
    \textbf{Anger} & $0.818$ & $0.333$ & $0.474$ & $27$ \\ 
    \textbf{Joy} & $0.812$ & $0.745$ & $0.777$ & $601$ \\ 
    \textbf{Neutral} & $0.903$ & $0.952$ & $0.927$ & $2,146$ \\ 
    \textbf{Sadness} & $0.864$ & $0.464$ & $0.604$ & $110$ \\ 
    \hline \hline
    \textbf{Micro AVG} & $0.885$ & $0.885$ & $0.885$ & $2,884$ \\ 
    \textbf{Macro AVG} & $0.849$ & $0.624$ & $0.695$ & $2,884$ \\ 
    \textbf{Weighted AVG} & $0.882$ & $0.885$ & $0.879$ & $2,884$ \\ 
    \bottomrule
    \end{tabular}
\end{table}

\section{Conclusion and Future work}

In the present work, we propose \textbf{FriendsBERT} and \textbf{ChatBERT} for the multi-utterance emotion recognition task on \textit{EmotionLines} dataset.
The proposed models are adapted from BERT~\cite{devlin2018bert} with three main improvement during the model training procedure, which are the causal utterance modeling mechanism, specific model pre-training, and adapt weighted loss. 
The causal utterance modeling takes the advantages of the sentence-level context information during model inference. The specific model pre-training helps to against the bias in different text domain. The weighted loss avoids our model to only predict on large size sample.
The effectiveness and generalizability of the proposed methods are demonstrated from the experiments.

In future work, we consider to include the conditional probabilistic constraint $P ({\rm Emo}_{B} | \hat{\rm Emo}_{A})$. Model should predict the emotion based on a certain understanding about context emotions. This might be more reasonable for guiding model than just predicting emotion of ${\rm Sentence}_B$ directly. In addition, due to the limitation of BERT input format, ambiguous number of input sentences is now becoming an important design requirement for our future work. Also, personality embedding development will be another future work of the emotion recognition. The personality embedding will be considered as sentence embedding injected into word embedding, and it seems this additional information can contribute some improvement potentially.


\bibliographystyle{named}
\bibliography{ijcai19}

\end{document}